\documentclass[letterpaper, 10 pt, conference]{ieeeconf}  
\IEEEoverridecommandlockouts
\usepackage{cite}
\usepackage{amsmath,amssymb,amsfonts}
\usepackage{algorithmic}
\usepackage{graphicx}
\graphicspath{{Figures/}}
\usepackage{textcomp}
\usepackage{subfigure}
\usepackage{verbatim}
\usepackage{xcolor}
\usepackage{multirow}
\usepackage{adjustbox}
\usepackage{textcomp}
\usepackage{epsfig}
\usepackage{import}
\usepackage{subfiles} 
 \usepackage{comment}
\usepackage{transparent}

\makeatletter
\newcommand\makebig[2]{%
	\@xp\newcommand\@xp*\csname#1\endcsname{\bBigg@{#2}}%
	\@xp\newcommand\@xp*\csname#1l\endcsname{\@xp\mathopen\csname#1\endcsname}%
	\@xp\newcommand\@xp*\csname#1r\endcsname{\@xp\mathclose\csname#1\endcsname}%
}
\makeatother

\makebig{biggg} {3.0}
\makebig{Biggg} {3.5}
\makebig{bigggg}{4.0}
\makebig{Bigggg}{4.5}

\def\BibTeX{{\rm B\kern-.05em{\sc i\kern-.025em b}\kern-.08em
    T\kern-.1667em\lower.7ex\hbox{E}\kern-.125emX}}
\begin{document}

\renewcommand{\arraystretch}{1.5}

\title{Six-degree-of-freedom Localization Under Multiple Permanent Magnets Actuation\\
}

\author{Tom\'as da Veiga, Giovanni Pittiglio, \textit{Member, IEEE}, Michael Brockdorff, James H. Chandler, \textit{Member, IEEE},\\ and Pietro Valdastri, \textit{Fellow, IEEE}
	\thanks{Research reported in this article was supported by the Engineering and Physical Sciences Research Council (EPSRC) under grants number EP/R045291/1 and EP/V009818/1, and by the European Research Council (ERC) under the European Union’s Horizon 2020 research and innovation programme (grant agreement No 818045). Any opinions, findings and conclusions, or recommendations expressed in this article are those of the authors and do not necessarily reflect the views of the EPSRC or the ERC.}
	\thanks{~Tom\'as~da~Veiga,
		Michael~Brockdorff,
		~James~H.~Chandler,
		~and~Pietro~Valdastri are with the STORM Lab, Institute of Autonomous Systems and Sensing (IRASS), School of Electronic and Electrical Engineering, University of Leeds, Leeds, UK.
		Email: \{\tt{eltgdv, elmbr, j.h.chandler, p.valdastri}\}@leeds.ac.uk}
	\thanks{Giovanni Pittiglio is with the Department of Cardiovascular Surgery, Boston Children’s Hospital, Harvard Medical School, Boston, MA 02115, USA. Email:
	{\tt giovanni.pittiglio@childrens.harvard.edu}}
}

\maketitle

\begin{abstract}
Localization of magnetically actuated medical robots is essential for accurate actuation, closed loop control and delivery of functionality. Despite extensive progress in the use of magnetic field and inertial measurements for pose estimation, these have been either under single external permanent magnet actuation or coil systems. With the advent of new magnetic actuation systems comprised of multiple external permanent magnets for increased control and manipulability, new localization techniques are necessary to account for and leverage the additional magnetic field sources. In this letter, we introduce a novel magnetic localization technique in the Special Euclidean Group SE(3) for multiple external permanent magnetic field actuation and control systems. The method relies on a milli-meter scale three-dimensional accelerometer and a three-dimensional magnetic field sensor and is able to estimate the full 6 degree-of-freedom pose without any prior pose information. We demonstrated the localization system with two external permanent magnets and achieved localization errors of 8.5 $\pm$ 2.4~mm in position norm and 3.7 $\pm$ 3.6$^{\circ}$ in orientation, across a cubic workspace with \textbf{20~cm} length.
\end{abstract}


\section{Introduction} \label{sec:intro}

Magnetically actuated medical robots (MAMR) have seen significant focus and development in recent decades due to their potential for miniaturization~\cite{hu2018small}, tether-less actuation~\cite{popek2016six} and high number of controllable degrees-of-freedom (DOFs)~\cite{salmanipour2018eight,pittiglio2022collaborative}. In fact, magnetically guided catheters have been used to treat cardiac arrhythmias since 2003~\cite{nelson2022magnetically,carpi2009stereotaxis}.

A key aspect in their actuation is pose estimation~\cite{bianchi2019localization,barducci2019adaptive}, enabling closed loop control and delivery of functionality~\cite{norton2019intelligent}. Imaging techniques have long been used for this purpose but are generally tied to limited resolution, harmful radiation exposure and need for additional hospital equipment~\cite{aziz2020medical,pane2022ultrasound,daguerre2022localization}. As such, methods based on magnetic field measurements have received significant attention, with magnetic tracking systems being widely available on the market. These, however, are not compatible with magnetic actuation systems due to distortions on the localization magnetic fields.

To address this issue, significant research on magnetic localization coupled with magnetic actuation systems has been done~\cite{khalil2019magnetic,popek2016six,son2018simultaneous,shao2019novel,taddese2018enhanced}. Several works have been based on magnetic field sensing arrays external to MAMR~\cite{micheal20222d,son2018simultaneous}. While advantageous from a miniaturization and internal power consumption point of view, these systems require calibration of large sensor arrays and have limited localization workspace dimensions. Internal sensing to the MAMR, on the other hand, does not suffer from workspace dimension restrictions. It requires, however, on-board power and heterogeneous localization magnetic fields, with 6-DOF localization having been shown for systems with a single external permanent magnet (EPM). \textcolor{black}{Internal sensing methods have been shown for endoscopic capsules, as well as magnetically guided catheters}~\cite{popek2016six,sperry2022six,taddese2018enhanced,fischer2022using}.

Over recent years the need for enhanced control and manipulability of MAMRs has led to the advent of actuation platforms based on multiple magnetic field sources (MMFS) such as multiple electromagnetic coils and multiple permanent magnets~\cite{kummer2010octomag,hoang2019independent,hong2020magnetic,pittiglio2022collaborative,ryan2017magnetic,stereotaxis_patent}. Some of these platforms have been cleared for human use such as Stereotaxis Genesis RMN\textsuperscript{\tiny\textregistered} based on two permanent magnets, and Magnetecs and Aeon Scientific based on multiple electromagnetic coils. 

Despite this progress, magnetic localization for such systems is lagging behind, with fluoroscopic imaging being currently used~\cite{nelson2022magnetically}. Unlike single magnetic field source systems where the singularity regions and localization limitations have been thoroughly investigated and solved for~\cite{taddese2018enhanced}, magnetic localization for MMFS systems suffers from additional challenges due to the superposition of the magnetic fields leading to configuration-specific singularity regions. Only recently, a 3D position localization system with internal magnetic field sensing was demonstrated for a multi-coil system, \textcolor{black}{for a 3~mm catheter}~\cite{fischer2022using}.

Furthermore, a common conundrum in 6-DOF magnetic localization with internal sensing is finding the rotation around gravity, due to the absence of the Earth's magnetic field measurement~\cite{mahony2008nonlinear}. This has been solved in the past by accurately initializing this missing rotation angle and tracking it with a gyroscope~\cite{pittiglio2020observability, di2016jacobian}. However, this is prone to errors over time, especially for slow moving systems where gyroscope data is not as sensitive. Additionally, if communication to the MAMR is lost, a new accurate initialization is needed, proving impossible mid medical intervention. More recently, Taddese et al.~\cite{taddese2018enhanced} fitted an auxiliary coil around a single EPM providing a second set of magnetic field measurements. This solves the missing rotation angle and is also able to eliminate the localization singularity plane when it comes to localization with respect to a single EPM. However, when MMFS are present in the workspace, that singularity plane ceases to exist due to the superposition of magnetic fields, and instead singularity regions are present depending on the relative pose of each EPM.

This paper introduces, for the first time, a 6-DOF magnetic localization method for systems with multiple EPMs without the need for any prior pose information. The method relies on measurements from an accelerometer and a single 3D magnetic field Hall effect sensor (HE), both internal to the MAMR. We analyze the effect that the number of EPMs in the workspace has on the full pose estimation; and demonstrate its performance in a two EPM magnetic actuation platform. Since adding an orthogonal coil is not able to solve for the singularity regions, in this work we do not consider it and instead solve for the missing rotation angle by using multiple magnetic field measurements at different EPM configurations. This works for static or quasi-static systems, with maximum MAMR velocity highly dependent on the actuation system and the magnetic field generated. This is the case for non-actuated parts of a larger system, such as the deployment point at the tip of an endoscope, or for MAMRs while the generated magnetic fields are sufficiently weak to induce actuation. Additionally, unlike common works in literature which parameterize the rotation matrix, in this work the full 6-DOF pose is estimated directly in the special euclidean group $SE(3)$. This avoids any singularities or non-unique representations of the orientation when using Euler angles or quaternions~\cite{mathavaraj2021se,mayhew2011quaternion,taddese2018enhanced}.

\section{Localization Strategy} \label{sec:loc-strategy}

\subsection{Problem Formulation}
\begin{figure}[h!]
	\centering
	\def\svgwidth{\columnwidth}
	\resizebox{.8\linewidth}{!}{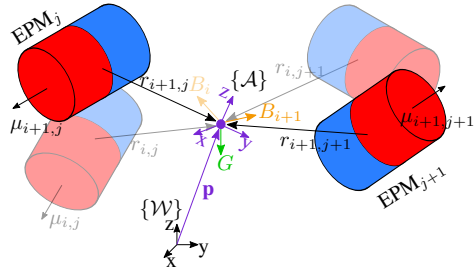}
	\caption{Representation of the world reference frame \{$\mathcal{W}$\} and MAMR reference frame \{$\mathcal{A}$\}, together with gravity vector $G$ in green, and magnetic field measurements $B_i$ in orange for \textit{m} EPMs. In purple is the state to estimate.}
	\label{fig:loc-alg}
\end{figure}

We consider finding the pose of a MAMR, with frame \{$\mathcal{A}$\} within our workspace  \{$\mathcal{W}$\} (see Fig.~\ref{fig:loc-alg}). Its position is denoted as $ \textbf{p} \in \mathbb{R}^3$ in \{$\mathcal{W}$\} and attitude as rotation matrix $ R \in SO(3)$ of the MAMR frame \{$\mathcal{A}$\} relative to \{$\mathcal{W}$\}. Additionally, the MAMR's linear velocity is denoted by $\textbf{V} \in \mathbb{R}^3$ expressed in \{$\mathcal{W}$\}. The MAMR's angular velocity expressed in \{$\mathcal{W}$\} relative to \{$\mathcal{A}$\} is represented by $\boldsymbol{\Omega} \in \mathbb{R}^3$.

We describe our state in the special euclidean group $SE(3)$, i.e. the group of homogenous transformations with entries in $\mathbb{R}^3$ associated with the Lie algebra, $\mathfrak{se}(3)$ of dimension 6. The main goal is to estimate the homogenous transformation matrix from the MAMR reference frame \{$\mathcal{A}$\} to the global frame \{$\mathcal{W}$\} (see Fig.~\ref{fig:loc-alg}).
\begin{equation*}
	T = ^\mathcal{W}T_\mathcal{A} : \{A\} \rightarrow \{W\}
\end{equation*}

Therefore, the dynamics model can be represented as
\begin{equation} \label{eq:SE3-system}
\dot{T} = T \begin{bmatrix}
(\boldsymbol{\Omega} + \textbf{b} + \boldsymbol{\delta})_\times & \textbf{V} \\
0 & 1
\end{bmatrix}
\end{equation}
with $(\boldsymbol{\Omega} + \textbf{b} + \boldsymbol{\delta})$ the measured angular velocity from the gyroscope including its bias $\textbf{b}$ and noise $\boldsymbol{\delta}$. Additionally, $(\cdot)_\times$ denotes the skew-symmetric matrix associated with the cross product by itself. 

\subsection{Measurement Model}
We consider our MAMR to be under $m$ EPMs actuation, and to be fitted with an accelerometer and a 3D HE sensor, providing two types of measurements: acceleration, and magnetic field vector.

Considering that gravitational acceleration ($\textbf{g}$) dominates over linear accelerations as per common approach in literature~\cite{mahony2008nonlinear,pittiglio2020observability}, the accelerometer measurement can be represented as (see Fig.~\ref{fig:loc-alg} in green)
\begin{equation}
	\textbf{G} = R^T\textbf{g}
\end{equation}
\textcolor{black}{where $R^T$ denotes the transpose of the MAMR's rotation matrix.}

The magnetic field vector generated by an EPM$_j$ (with $j = 1, ..., m$) can be modeled as a dipole
\begin{equation} \label{eq:dipole_model}
\textbf{B}_{j} :=  \textbf{B}(\mu_{j},\textbf{r}_{j}) = \frac{\mu_0 |\mu_{j}|}{4 \pi |\textbf{r}_{j}|^3} \left(3 \hat{\textbf{r}}_{j} \hat{\textbf{r}}_{j}^T - I \right)\hat \mu_{j}
\end{equation}
with $\textbf{r}_{j}$ the distance between \{$\mathcal{A}$\} and EPM$_j$, and $\mu_{j}$ the EPM's magnetic moment in \{$\mathcal{A}$\}. This assumption is valid for far-enough distances from the EPMs and is commonly employed in other magnetic localization works~\cite{petruska2012optimal,taddese2018enhanced}. Assuming that there are no metal objects in the workspace, the measured magnetic field $\textbf{B}$ equals the sum of the magnetic fields generated by each EPM.
\begin{equation} \label{eq:mag_model}
	\textbf{B} = \sum_{j = 1}^{m} \textbf{B}_{j}
\end{equation}

Given the absence of the Earth's magnetic field measurement, a minimum of two magnetic field measurements for different configurations of the $m$ EPMs are needed for observability (see Fig.~\ref{fig:loc-alg} in orange, and Section~\ref{sec:observability}). This is a valid assumption for systems where the magnetic field changes much quicker than the MAMR's pose, such as static or quasi-static systems. This being so, assuming null mean Gaussian measurement noises~\cite{mahony2008nonlinear}, the measurement model can be expressed as follows. \textcolor{black}{In addition to $n$ measurements of the magnetic field, their norm $\|\textbf{B}_i\|$ was also included. Unlike the full magnetic field measurement vector, which contains information on both position and orientation, the magnetic field norm is dependent only on the MAMR's position. When multiple measurements are present, the addition of the magnetic field norm increased convergence speed.}
\begin{equation} \label{eq:meas_model}
	\textbf{h} = \begin{bmatrix}
	\|\textbf{B}_1\| \\
	\vdots \\
	\| \textbf{B}_i \| \\
	\textbf{B}_1 \\
	\vdots \\
	\textbf{B}_i \\
	\textbf{G}
	\end{bmatrix}, \qquad i = 2, \ldots, n
\end{equation}

\subsection{Extended Kalman Filter}
Extended Kalman Filters (EKF) in $SO(3)$ and $SE(3)$ have been widely used and proved effective~\cite{pittiglio2020observability,mathavaraj2021se}. \textcolor{black}{For the sake of summary, only the EKF equations are explicitly described here. Further detail on the formulation of EKF can be found in~\cite{pittiglio2020observability} and~\cite{barfoot2017state}}.

The discrete dynamics of the estimated state can be described as 
\begin{align} \label{eq:dynamics}
	\hat{T}_{k+1} = \hat{T}_k \text{exp}(\text{K}_k\tilde{\textbf{y}}_kt) \\
	\tilde{\textbf{y}}_k = \textbf{y}_k - \textcolor{black}{\textbf{h}}(\hat{T}_k)
\end{align}
with time-step $k = 0, t, 2t, ...$, $K_k$ the gain defined by the standard EKF prediction and update steps below, $\text{exp}(\cdot)$ the exponential map of $SE(3)$, \textcolor{black}{$\textbf{h}$ the measurement model defined in eq.(\ref{eq:meas_model}), and $\textbf{y}_k$ the sensors' outputs in the measurement model format, i.e. the norm of the magnetic field, followed by the magnetic field and gravity.}

\subsubsection{Prediction} \textcolor{black}{This step sees the propagation of the state covariance matrix $P_k \in \mathbb{R}^{6 \times 6}$ as}
\begin{equation*}
P_k = F_k\overline{P}_{k-1}F_k^T + G_kQ_nG_k^T
\end{equation*}	
\textcolor{black}{with $P_k = \text{diag}(P_{k_{p}}, P_{k_{R}})$, where $P_{k_{p}}$ and $P_{k_{R}}$ denote the state covariance matrix of the position and orientation respectively. Additionally, input noise is considered as a null-mean Gaussian distribution with constant covariance $Q_n \in \mathbb{R}^{6\times6}$. Lastly, $F_k = \text{exp}(A_k t)$ and $G_k = T_k \frac{\partial}{\partial A_k}\text{exp}(A_k t)$, with $A_k$ defined by the Lie algebra as matrix $A_k = [\Omega_\times \quad V; 0 \quad 0]$}.

\subsubsection{Update} The second step sees the computation of the gain $K_k$ used in the update of the state as shown in eq.~(\ref{eq:dynamics}) through
\begin{align*}
	S_k & = H_kP_kH_k^T + R_n \\
	K_k & = P_kH_k^TS_k^{-1} \\
	\overline{P}_k & = P_k - K_kS_kK_k^T \\
\end{align*}
\vspace{0.5cm}
where $H_k = \frac{\partial \textcolor{black}{\textbf{h}_k}}{\partial T_k} $. Additionally, measurement noise is considered as a null-mean Gaussian distribution with constant covariance matrix $R_n \in \mathbb{R}^{m\times m}$ - $\textcolor{black}{\textbf{h}} \in \mathbb{R}^m$.

\subsection{Error metrics: } The observer's performance was assessed through two different error metrics: one for the estimation of the MAMR's position and one for the MAMR's attitude.

\begin{IEEEeqnarray*}{rCl}
	\text{e}_p = \|\textbf{p} - \hat{\textbf{p}}\|\\
	\text{e}_R = \text{tr}(I - \hat{R}^TR) \IEEEyesnumber
\end{IEEEeqnarray*}

\section{Simulation} \label{sec:simulation}

\vspace{-0.15cm}
To infer the stability and performance of the observer, first, an observability analysis on the system was done to assess the \textcolor{black}{minimum number of magnetic field measurements for observability. Second, the impact the number of EPMs $m$ and the maximum number of magnetic field measurements $n$ in the measurement model (see eq.~(\ref{eq:mag_model}) and~(\ref{eq:meas_model})) have on the stability of the observer was analyzed. Lastly, the observer was run within a simulated environment to infer the EKF's performance and expected convergence time. EKF covariance matrices $P_0$, $Q_n$ and $R_n$ were tuned in this step.}

\begin{figure}[!ht]
	\centering
	\includegraphics[scale=0.2]{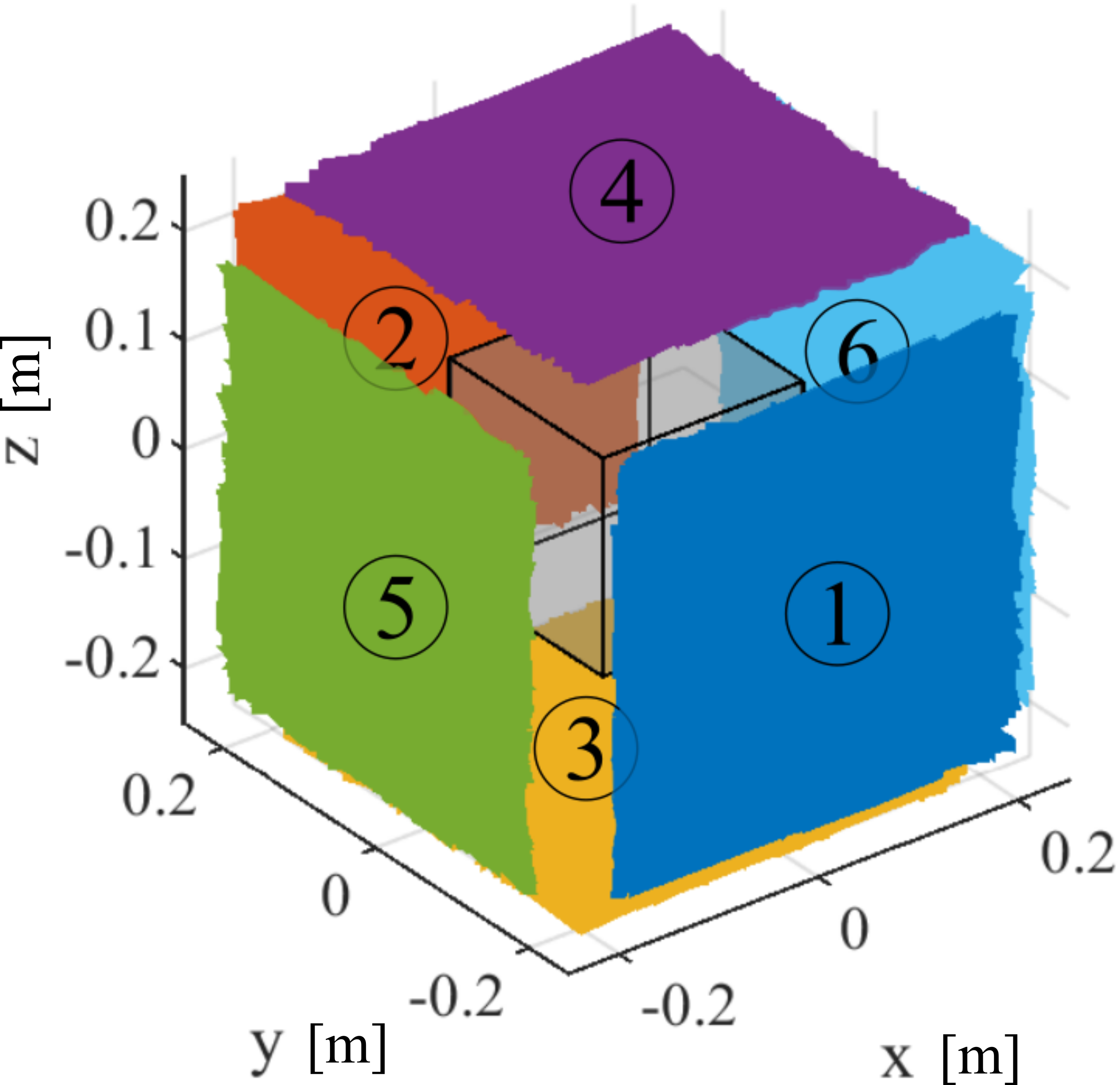}
	\caption{Planes covered by the generated EPM paths. Each EPM is constrained to a plane 15~cm from the workspace edge.}
	\label{fig:EPMs_paths}
\end{figure}

The number of EPMs was varied between one and six. \textcolor{black}{Given that EPMs are used for actuation, localization should not rely on a specific EPM motion. Therefore, random motion paths were generated for each EPM.} Additionally, each EPM was constrained to a plane 15~cm away from the workspace edge, as seen in Fig.~\ref{fig:EPMs_paths}.
\vspace{-0.4cm}
\subsection{Observability Analysis} \label{sec:observability}
\textcolor{black}{To assess the minimum number of magnetic field measurements $n$ needed for observability}, an observability analysis was performed for system in eq.~(\ref{eq:SE3-system}) with measurement model in eq.~(\ref{eq:meas_model}). Local weak observability of a non-linear system is defined by the following codistribution being full rank, i.e $\text{rank}(\nabla_T \mathcal{O}) = 6$.
\begin{equation}
	\nabla_T \mathcal{O} = \text{span}(\{\nabla_T \mathcal{L}_{\dot{T}}^i\textbf{ h}, i \in \mathcal{N}^+ \cup 0\})
\end{equation}
where $\mathcal{L}_{\dot{T}}^i \textbf{h}$ defines the $i$th-order Lie derivative of $\textbf{h}$ with respect to the state $T$. \textcolor{black}{Further details on the notation and derivation of an observability analysis can be found in~\cite{pittiglio2020observability}}. In this work, we consider the first order derivative only, and so, this codistribution can be expanded as
\begin{equation}
\nabla_T \mathcal{O} = \left[  \nabla_p \mathcal{O} \quad \nabla_R \mathcal{O} \right ] =  \begin{bmatrix}
\nabla_p \mathcal{O}_{\|\textbf{B}_1\|} & \nabla_R \mathcal{O}_{\|\textbf{B}_1\|} \\
\vdots & \vdots\\
\nabla_p \mathcal{O}_{\|\textbf{B}_n\|} & \nabla_R \mathcal{O}_{\|\textbf{B}_n\|} \\
\nabla_p \mathcal{O}_{\textbf{B}_1} & \nabla_R \mathcal{O}_{\textbf{B}_1} \\
\vdots & \vdots\\
\nabla_p \mathcal{O}_{\textbf{B}_n} & \nabla_R \mathcal{O}_{\textbf{B}_n} \\
\nabla_p \mathcal{O}_{\textbf{G}} & \nabla_R \mathcal{O}_{\textbf{G}}\\
\end{bmatrix}
\end{equation}
making explicit the two components of the state, position and orientation, and the different types of measurement.

As shown in~\cite{pittiglio2020observability}, $\nabla_R \mathcal{O}$ represents the Lie derivative with respect to the orientation. Since the norm of the magnetic field has no orientation information, $\nabla_R \mathcal{O}_{\|\textbf{B}_i\|}$ is equal to zero. 
\begin{equation}
\nabla_R \mathcal{O}_{\|\textbf{B}_i\|} = 0_{1 \times 3}
\end{equation}
\begin{equation}
\nabla_R \mathcal{O}_{\textbf{B}_i} = \begin{bmatrix}
0 & -R_{:,3} \cdot \textbf{B}_i & R_{:,2} \cdot \textbf{B}_i \\
R_{:,3} \cdot \textbf{B}_i & 0 & -R_{:,1} \cdot \textbf{B}_i \\
- R_{:,2} \cdot \textbf{B}_i & R_{:.1} \cdot \textbf{B}_i & 0 \\
\end{bmatrix}
\end{equation}
\begin{equation}
\nabla_R \mathcal{O}_{G} = \begin{bmatrix}
0 & R_{33} & -R_{32} \\
-R_{33} & 0 & R_{31} \\
R_{32} & -R_{31} & 0
\end{bmatrix}
\end{equation}

$\nabla_p \mathcal{O}$ represents the Lie derivative with respect to the position, and can be expressed as follows. Given that IMU measurements only contain information regarding orientation, $\nabla_p \mathcal{O}_{G}$ is equal to zero. 
\begin{equation}
\nabla_p \mathcal{O}_{\|\textbf{B}_i\|} = \begin{bmatrix}
\frac{\partial \|\textbf{B}_i\|}{\partial x} & \frac{\partial \|\textbf{B}_i\|}{\partial y} & \frac{\partial \|\textbf{B}_i\|}{\partial z}
\end{bmatrix}
\end{equation}
\begin{equation}
\nabla_p \mathcal{O}_{\textbf{B}_n} = \begin{bmatrix}
\frac{\partial \textbf{B}_{i_x}}{\partial x} & \frac{\partial \textbf{B}_{i_x}}{\partial y} & \frac{\partial \textbf{B}_{i_x}}{\partial z} \\
\frac{\partial \textbf{B}_{i_y}}{\partial x} & \frac{\partial \textbf{\textbf{B}}_{i_y}}{\partial y} & \frac{\partial \textbf{\textbf{B}}_{i_y}}{\partial z} \\
\frac{\partial \textbf{B}_{i_z}}{\partial x} & \frac{\partial \textbf{B}_{i_z}}{\partial y} & \frac{\partial \textbf{B}_{i_z}}{\partial z} \\
\end{bmatrix}
\end{equation}
\begin{equation}
\nabla_p \mathcal{O}_{G} = 0_{3 \times 3}
\end{equation}

Looking at the full observability matrix  $\nabla_T \mathcal{O}$, we see that for when $n=1$, $\text{rank}(\nabla_T \mathcal{O}) = 5$ making the system not observable.  In fact, a single configuration of the EPMs and its respective magnetic field $\textbf{B}_i$ together with its norm and $\textbf{G}$ are not enough to solve the full 6-DOF pose. This can intuitively be inferred as the gravity vector measurement is able to provide 2-modes of the orientation, with the rotation around its own axis, i.e. rotation around gravity, missing~\cite{pittiglio2020observability}. Since the magnetic field vector and its norm are not linearly independent, only three of the remaining 4 modes of the pose dynamics can be solved for. Therefore, without any prior pose information, a minimum of 2 measurements of magnetic field are necessary in order to make the system observable and estimate the full 6-DOF pose. Additional measurements of the magnetic field can be taken for different EPM configurations.

\subsection{\textcolor{black}{Magnetic Analysis}}
\textcolor{black}{Having shown that a minimum of two magnetic field measurements for different EPM configurations are needed for observability, the effect this number ($ 2  \leqslant n  \leqslant 100$) has on the stability of the observer is further inferred. Additionally, the effect the number of EPMs ($ 1  \leqslant m  \leqslant 6$) in the workspace has on the stability was also analyzed. This was done by taking the condition number $N_c$ across multiple planes of the workspace for the different cases.} The condition number is defined as the ratio between the maximum and minimum singular values of $\nabla_T \mathcal{O}$, and, as such, lower values indicate a better conditioned system. 
\begin{figure}[!h]
	\centering
	\includegraphics[scale=0.8]{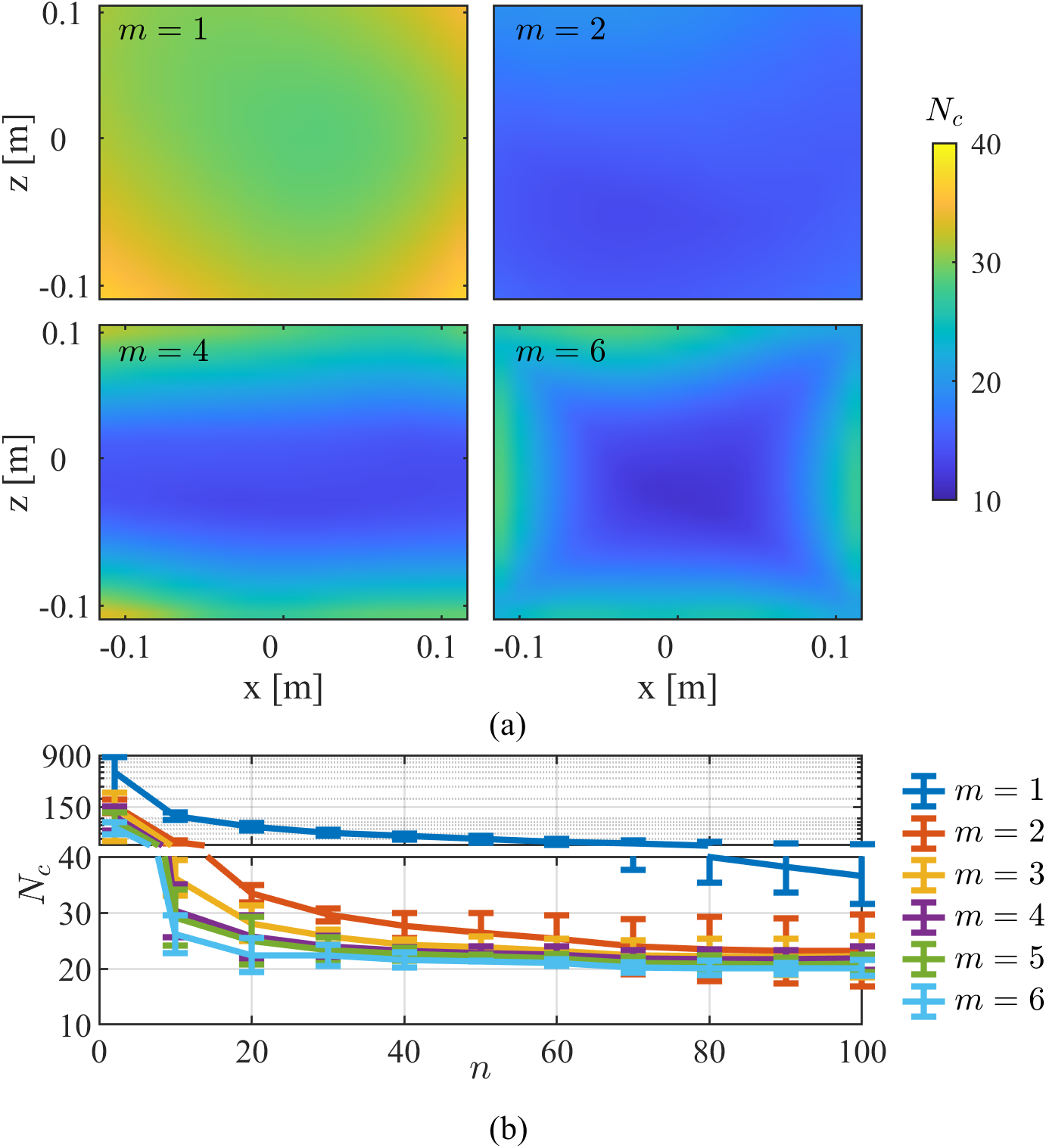}
	\caption{System's condition number $N_c$ for different numbers of EPMs $m$ and different number of EPM configurations $n$ in the model. (a) Shows the condition number $N_c$ across the XZ plane of the workspace for one, two, four and six EPMs, when $n = 100$. (b) Plot showing how the condition number $N_c$ changes with higher number of EPM configurations in the measurement model for each number of EPMs.}
	\label{fig:cond_num}
\end{figure}
Fig.~\ref{fig:cond_num}(a) shows $N_c$ across the XZ plane ($y = 0$) for $n=100$ and for one, two, four, and six EPMs in the workspace, respectively. Fig.~\ref{fig:cond_num}(b) plots how $N_c$ changes when multiple EPM configurations $n$ are added to the measurement model, for each number of EPMs. $N_c$ was computed at three planes of the workspace XY ($z=0$), XZ ($y=0$, represented in (a)), and YZ ($x=0$). As we can see, there is a significant difference between a single EPM $m=1$ and multiple EPMs $ m \geqslant 2$, with $ m \geqslant 2$ having significantly lower  $N_c$ for any number of EPM configurations $n$. This is due to the fact that when multiple EPMs are present in the workspace, the resulting magnetic field becomes considerably less trivial, reducing the number of possible solutions for a specific measured magnetic field. However, there is no significant difference for when $m$ increases beyond two. Additionally, $N_c$ lowers as more EPM configurations $n$ are added to the measurement model. However, a plateau is reached at around $n = 20$, as more EPM configurations do not lower $N_c$. 

\subsection{Simulated Observer} \label{sec:sim_obser}
To further predict the performance of the EKF, the observer was ran with the MAMR fixed at 100 different randomly generated poses across the workspace. Convergence was deemed achieved once position error was below 5~mm in all axis, and the trace of the orientation error under 0.1, both for over 150 consecutive time-steps. Since the number of EPM configurations $n$ in the measurement model affects the EKF's frequency due to robot movement and data acquisition time, rather than assessing speed through EKF iterations $k$, speed was assessed by the total number of different EPM configurations needed until convergence was reached, $ n \cdot k$. The results were averaged across all 100 tested MAMR poses for each number of EPMs and EPM configurations.

\begin{figure}[!ht]
	\centering
	\includegraphics[scale = 0.8]{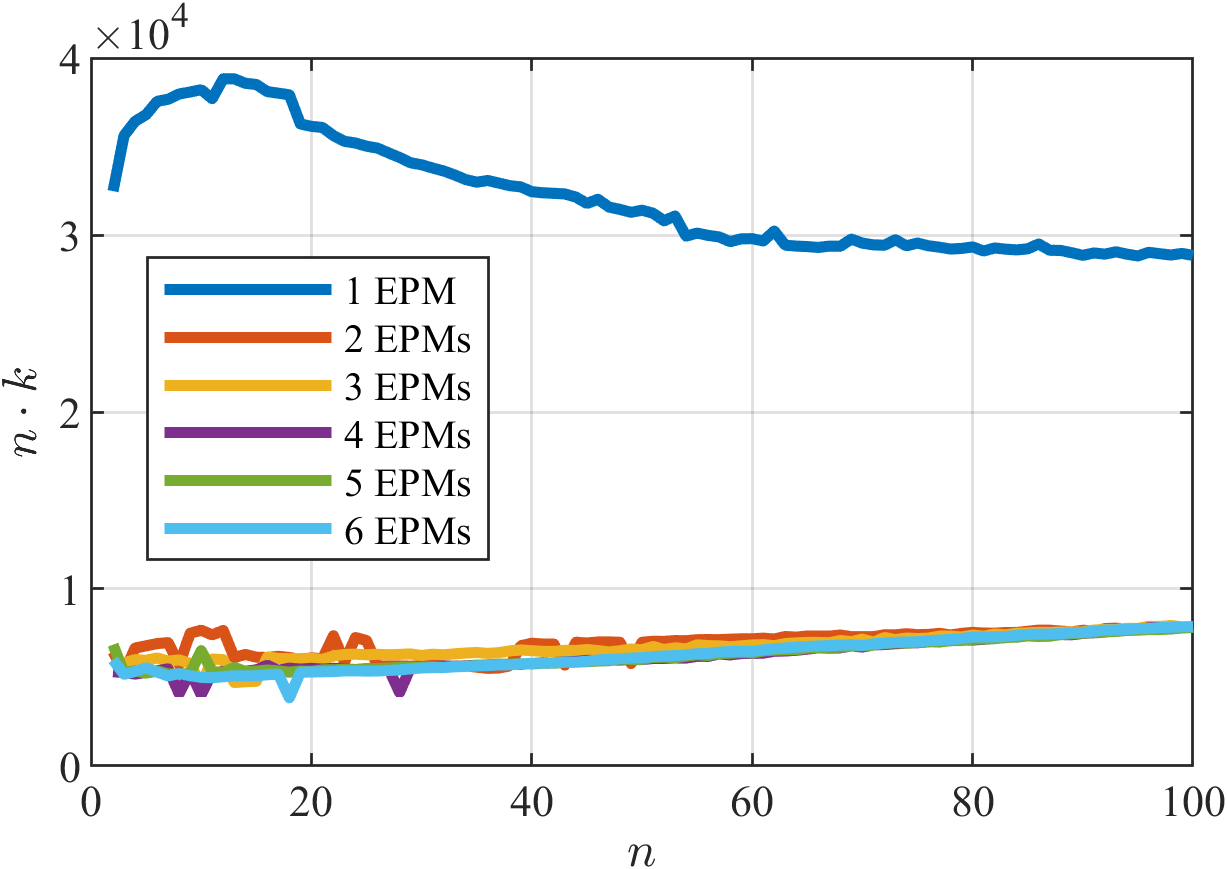}
	\caption{Effect that multiple EPMs and the number of EPM configurations in the measurement model $n$ have on convergence speed $n \cdot k$. Convergence was achieved once errors in position were below 5~mm across all axis, and the trace of the orientation error $e_R$ below 0.1, for over 150 consecutive time-steps.}
	\label{fig:sim_results}
\end{figure}

Fig.~\ref{fig:sim_results} plots the results obtained. As expected from the previous condition number analysis, there is a clear distinction between a single EPM and multiple EPMs. Multiple EPMs lead to a much faster convergence needing a significantly lower total number of EPM configurations. However, the difference between two and six EPMs is marginal. Additionally, the higher the number of EPM configurations $n$ in the measurement model the faster the convergence for a single EPM, as the associated $N_c$ gets lower. However, with multiple EPMs this effect is not as noticeable, with $n$ around $20$ leading to a faster convergence. 

\begin{figure}[!ht]
	\centering
	\includegraphics[scale=0.8]{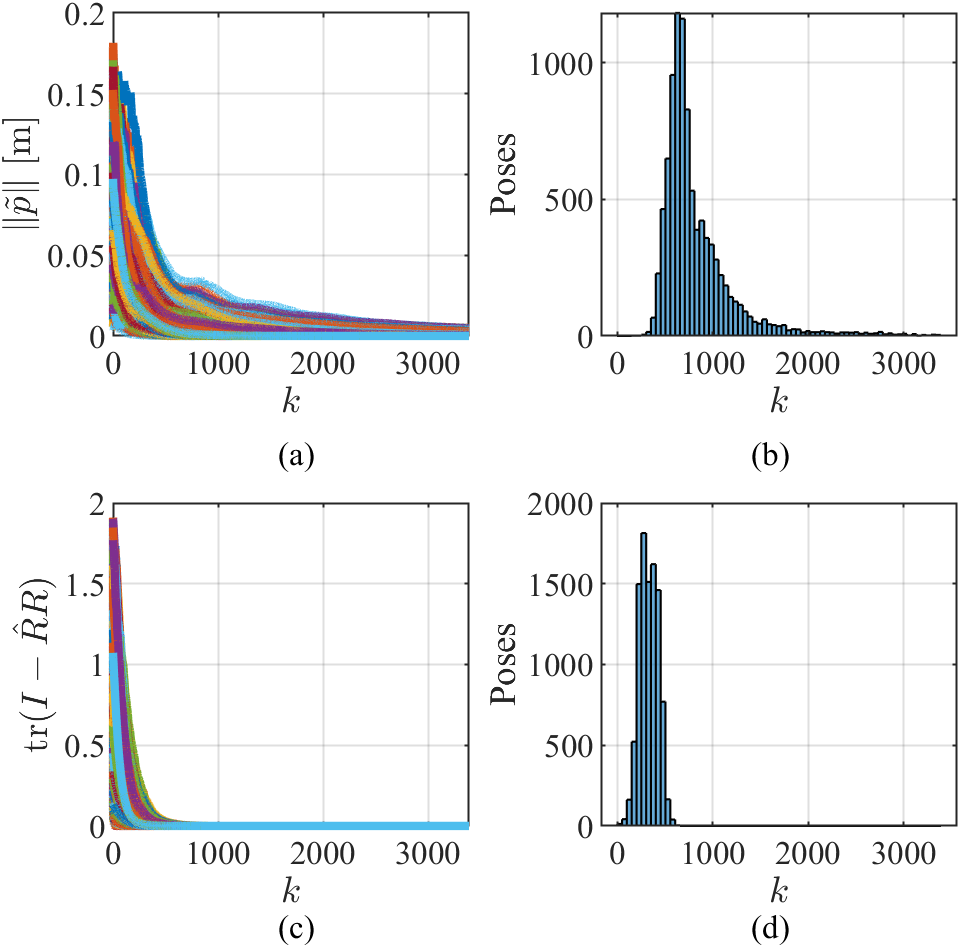}
	\caption{Simulation errors for 10,000 random poses across the workspace over EKF iterations, with 2 EPMs and 20 EPM configurations in the measurement model. (a) Norm of the position error, (b) Histogram showing the distribution of convergence in position, (c) Error in orientation, (d) Histogram showing the distribution of convergence in orientation.}
	\label{fig:sim_results_10k}
\end{figure}

Given these results, we consider from this point forward the case for which $m = 2$ and $n= 20$, i.e. there are two EPMs in the workspace, and the measurement model is comprised of 20 different EPM configurations. To further assess the localization performance for these conditions, a simulation was ran for 10,000 different random MAMR poses across the workspace. Fig.~\ref{fig:sim_results_10k}(a),(c) shows the error in position e$_p$ and orientation e$_R$ over time for all tested poses. As we can see, the observer converged for all tested poses, with 95.0\% of tested poses with norm position errors below 1~mm at finish. Additionally, as the histograms show, convergence in orientation is achieved faster than position, with 100\% of the poses having converged fully in orientation before 1000 iterations (see Fig.~\ref{fig:sim_results_10k}(d)).

\vspace{-0.3cm}

\section{Experimental Setup} \label{sec:exp-setup}

To evaluate the proposed localization system performance, a sensing platform was developed and tested with a 2-EPM system. 

The sensor board was composed by a 3D IMU (LSM6DS3, STMicroelectronics, Switzerland. Accelerometer sensing range $\pm2$g, Sensitivity $0.061$mg/LSB$_{16}$, Footprint $2.5\times3\times0.83$~mm) and a 3D HE (MLX90395, Melexis, Belgium. Sensing range $\pm 50$~mT; Sensitivity $2.5~\mu$T/LSB$_{16}$, Footprint $3\times3\times0.9$~mm). \textcolor{black}{The sensors used were chosen due to their dimensions, sensitivity and sensing range, allowing their use in embedded devices of the millimeter scale under high magnetic fields.} The sensors were interfaced with a Raspberry Pi 4B through I$^2$C protocol. The HE sensor was calibrated by placing it in the center of a 1D Helmholtz coil (DXHC10-200, Dexing Magnet Tech. Co., Ltd, Xiamen, China) under known magnetic field vectors.

\begin{figure}[!ht]
	\centering
	 \resizebox{.8\linewidth}{!}{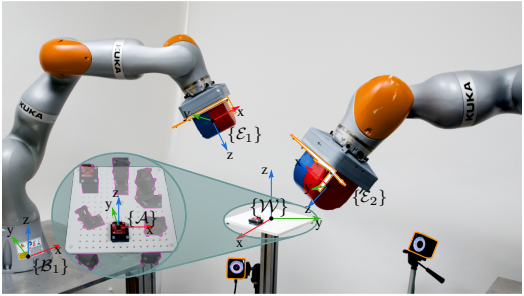}
	\caption{Experimental setup, comprised of two robotic arms with EPM at the end-effectors, Optical Tracking system, and sensor board.}
	\label{fig:setup}
\end{figure}

The dual EPM platform (dEPM) was used~\cite{pittiglio2022collaborative,pittiglio2022patient}, consisting on two KUKA LBR iiwa14 robots (KUKA, Germany), each manipulating one EPM (cylindrical permanent magnet with diameter and lenght of 101.6~mm and axial magnetization of 970.1~Am$^2$ (Grade N52)) (see Fig.~\ref{fig:setup}).

To fully assess the localization performance across the dEPM large workspace, a 3D printed plate (20-by-20~cm) was placed in between the two robots, delimiting the localization workspace in two dimensions. The sensor board was attached to 3D printed holders of various heights and orientations, which were in turn attached to the plate, allowing full variation of position and orientation.

Additionally, ground truth data was collected via a 4-camera optical tracking system (OptiTrack, Prime 13, NaturalPoint, Inc., USA, with submilimeter accuracy). With optical markers attached to the end-effectors of both robots, to the workspace plate and to the sensor board, the relative pose of each robotic arm base and the sensor board with respect to \{$\mathcal{W}$\} was found before each experiment (see Fig.~\ref{fig:setup}). While the EPMs were in motion, their poses were determined by reading the robotic arms joints and computing the inverse kinematics. This ensures a more accurate tracking of the motion of the EPMs since the markers may be blocked from the field of view during the motion. 

Finally, the Raspberry Pi, the robotic arms, and the optical tracking system were all connected using ROS. Data from the robotic arms encoders and sensors was collected at a rate of 50Hz. Given the inclusion of 20 EPM configurations in the measurement model, the EKF was ran at $50/20 = 2.5$~Hz. The EKF parameters used are shown in Table~\ref{tb:EKF_param}. These were determined by the simulation step in Section~\ref{sec:sim_obser} and the sensors used. \textcolor{black}{Additionally, the state was initialized at the origin of the workspace, $T_0 = I$.}
\begin{table}[h!]
	\caption{EKF Covariance Matrices} \label{tb:EKF_param}
	\centering
	\begin{tabular}{c|c|}
		\cline{2-2}
		\textbf{} & \textbf{EKF} \\ \hline
		\multicolumn{1}{|l|}{\textbf{State}} & $P_0 = \text{diag}(10^{-4}, 10^{-4})$ \\ \hline
		\multicolumn{1}{|l|}{\textbf{Input}} & $Q_n = \text{diag}(10^{-5}, 10^{-3})$ \\ \hline
		\multicolumn{1}{|l|}{\textbf{Measurement $\|\textbf{B}\|$}} & $R_{n_{\|B\|}} = 10^{-8}$ \\ \hline
		\multicolumn{1}{|l|}{\textbf{Measurement $\textbf{B}$}} & $R_{n_B} = \text{diag}(10^{-8},10^{-8},20^{-8})$ \\ \hline
		\multicolumn{1}{|l|}{\textbf{Measurement $\textbf{G}$}} & $R_{n_G} = 10^{-6}I$ \\ \hline
	\end{tabular}
\end{table}

\section{Results} \label{sec:results}

\begin{figure}[!ht]
	\centering
	\includegraphics[scale=0.8]{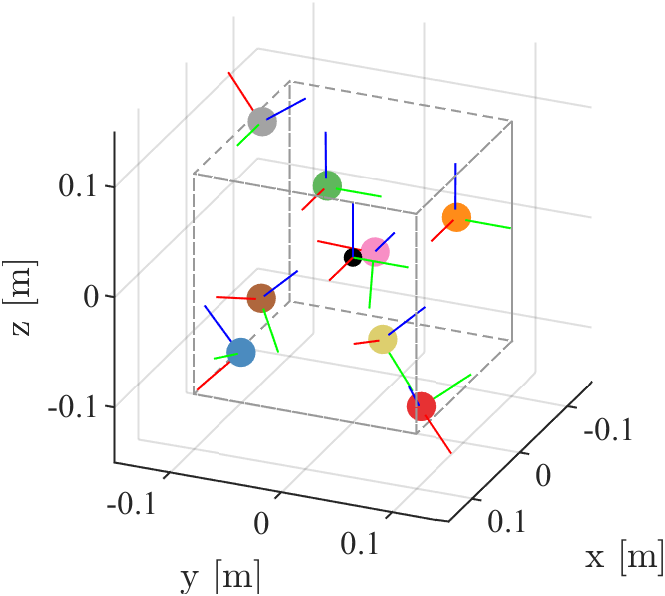}
	\caption{Tested poses across the workspace.}
	\label{fig:tested_poses}
\end{figure}

The localization algorithm was tested for eight different poses across the workspace (see Fig.~\ref{fig:tested_poses}). Each pose was tested twice, with the EPMs doing a different random motion each time composed of 200 different poses.

\begin{figure*}[!ht]
	\vspace{0.5cm}
	\centering
	\includegraphics[scale=0.8]{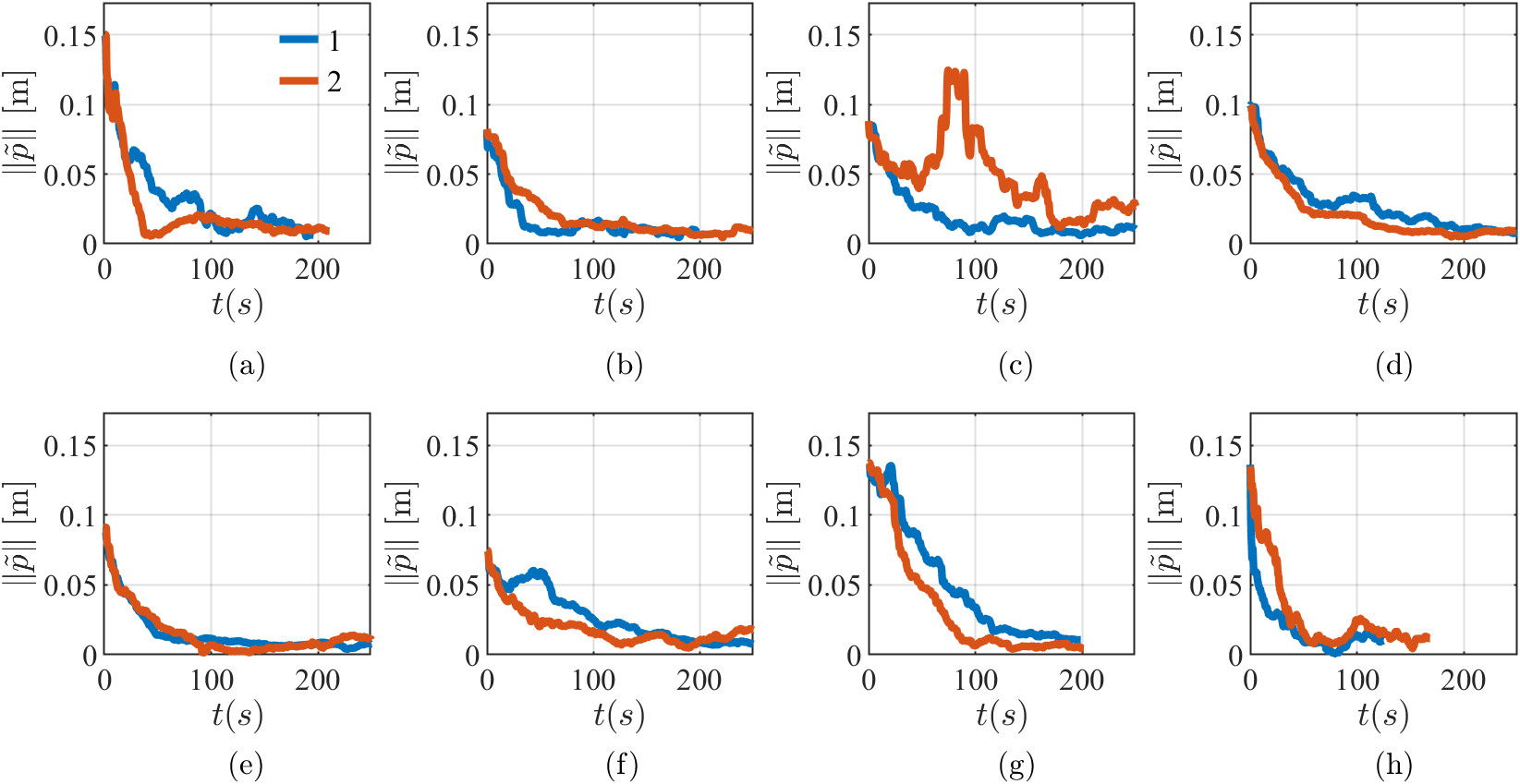}
	\caption{Error in position estimation for the ten tested poses across the workspace. Two repeats for each pose were performed.}
	\label{fig:pos_results}
\end{figure*}

\begin{figure*}[!ht]
	\centering
	\includegraphics[scale=0.8]{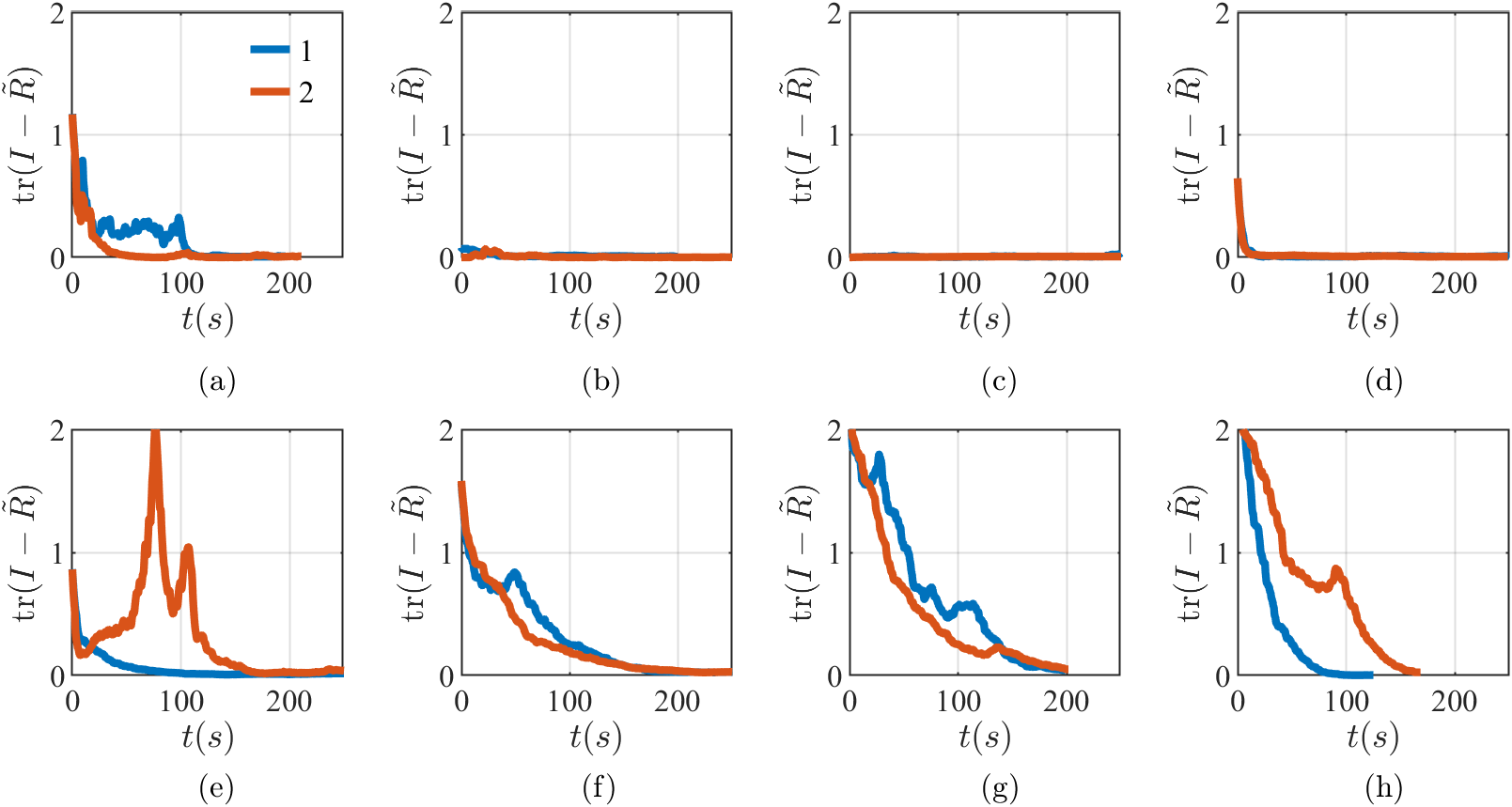}
	\caption{Error in orientation estimation for the ten tested poses across the workspace. Two repeats for each pose were performed.}
	\label{fig:ori_results}
\end{figure*}

Fig.~\ref{fig:pos_results} and~\ref{fig:ori_results} depict the error in position e$_p$ and orientation e$_R$ respectively, for each tested pose and repeat. The observer converged to the right solution for all tested poses with an average error of 8.5 $\pm$ 2.4 mm in position norm - with 4.14 $\pm$ 3.0 mm along the X axis, 4.13 $\pm$ 3.0 mm on the Y axis, and 3.44 $\pm$ 2.5 mm along the Z axis - and $0.032 \pm 0.027$ in orientation trace error, i.e. 3.7 $\pm$ 3.6$^{\circ}$.

However, as Fig.~\ref{fig:pos_results} and~\ref{fig:ori_results} show, there is significant variation in convergence speed and stability of the solution across repeats for the same pose. Given that the only difference between repeats is the EPMs motion, and therefore, the magnetic field measured by the sensors, the path each EPM takes and their combination have a big impact on the algorithm performance. This seems to be more significant for the estimation of the position than for the orientation, given that position estimation relies exclusively on magnetic field measurements. Fig.~\ref{fig:pos_results}(c) illustrates this effect very clearly, where for repeat 1 the algorithm converged to the right solution only to start diverging towards the end, and repeat 2 took longer to converge than all other cases. Unlike localization with a single EPM where the localization singularity plane is well defined and known, when multiple EPMs are present in the workspace, their relative pose dictates whether there are singularity regions and where they are. \textcolor{black}{Since the EPMs are travelling random paths}, it is possible that at times the sensors were located in a singularity region. Given the presence of multiple EPM configurations at each iteration of the observer, this does not seem to impact convergence but rather convergence speed. If the measurement model contained only a single configuration of EPMs, ideal for fast moving MAMR, these singularity conditions would need to be well defined and avoided.

\begin{figure}[!ht]
	\centering
	\includegraphics[scale=0.8]{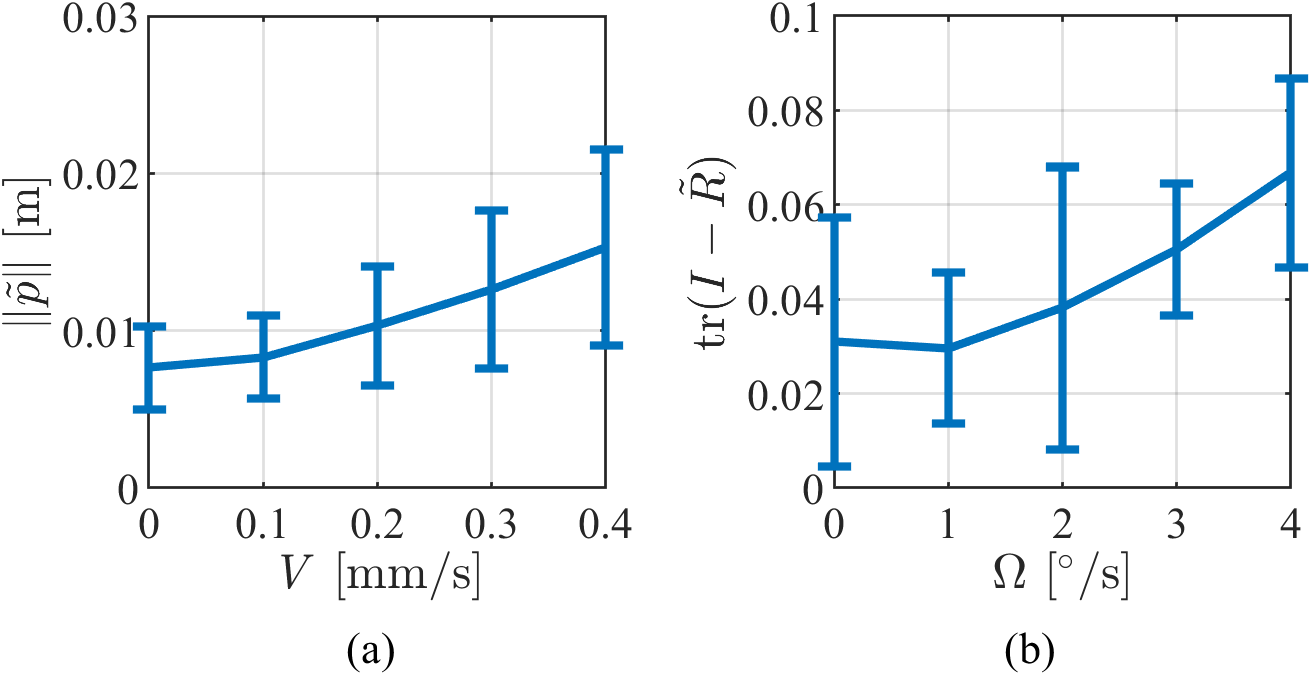}
	\caption{Error in (a) position and (b) orientation estimation for different linear and angular MAMR velocities.}
	\vspace{-0.5cm}
	\label{fig:speed_results}
	
\end{figure}

To test the observer's behavior in non-static conditions two different scenarios were tested. First, to address periodic motions such as breathing, linear and angular velocities were given to the world reference frame \{$\mathcal{W}$\} in the previous set of experiments as to mimic MAMR motion. The obtained results are shown in Fig.~\ref{fig:speed_results}. Linear velocities of up to 0.1~mm/s and angular velocities up to 2 $^\circ$/s produced marginal differences when compared to the static cases. Velocities above these values had significant impact on the results. Second, the observer's robustness for occasional spike movements such as coughing was tested. Spikes of 5~cm of up to 4~seconds, and spikes of 10~cm of up to 2~seconds did not produce significant changes in results. Longer spike times made the results unreliable. These values, however, are highly dependent on the platform. In this case, the robotic arms were operating at 30\% of their full speed for safety reasons. Increasing this speed, and/or including less EPM configurations in the measurement model, would allow faster MAMR speeds and longer spike motions.

\section{Conclusions} \label{sec:conclusions}

In this letter, a 6-DOF localization strategy without any prior pose information for actuation systems under multiple EPMs was presented. \textcolor{black}{The method relies on the measurements from a 3D accelerometer and a 3D HE sensor. These sensors are low-cost and widely available. Additionally, their small footprint makes them easily embedded in small-scale medical robots. In fact, magnetic localization based on these sensors has long been in use in medical robots, ranging from catheters~\cite{fischer2022using} to endoscopic capsules~\cite{taddese2018enhanced}. However, as new platforms based on multiple EPMs emerge for the control and actuation of magnetically actuated continuum robots for endoluminal procedures, localization techniques that take into account multiple magnetic field sources are needed. The internal placement of the sensors to the MAMR should be carefully designed to better offset any internal magnetic field measurements from the sensor. This will ensure accurate external magnetic field readings}.

\textcolor{black}{Unlike previous work that shows localization with respect to a single EPM, in this work we developed a localization technique under multiple EPM control}. We showed that, when compared to a single EPM, multiple EPMs lead to faster convergence speeds. The method was tested across a $8000~\text{cm}^3$ workspace, with average errors of 8.5 $\pm$ 2.4 mm in position norm and $0.032 \pm 0.027$ in orientation trace error. \textcolor{black}{This localization technique can thus be applied to endoscopic capsules, or magnetically guided catheters, which are under MMFS control or in close proximity to additional magnetic field sources.}

\textcolor{black}{In this work, the EPMs were moved randomly around the workspace, as their movement should be mainly optimized for actuation}. However, this was shown to lead to localization singularity regions and varying results when it comes to convergence speed and error. \textcolor{black}{Optimizing the EPM paths for both actuation and localization for active sensing will allow for reliable simultaneous localization and actuation under multiple EPM control. This could be achieved by analysing each specific EPM configuration required for actuation, and finding an alternative whenever such a configuration leads to non-observability. Additionally, this would allow a reduction in the number of measurements needed per time step, increasing the state estimation update rate and convergence speed.}

Lastly, \textcolor{black}{the speed at which the EPMs are moving is crucial for the convergence speed of the observer, as well as, the MAMR's speed}. Due to multiple instances of the EPM configurations present in the measurement model, a significant change in magnetic field should be captured across different EPM configurations. With the robotic arms moving at 30\%\ of their full speed and 20 EPM configurations per iteration, the observer was running at 2.5~Hz allowing MAMR's speeds of up to 0.2~mm/s. \textcolor{black}{The robotic arms speed was constrained for safety reasons due to the random motion travelled. It is expected that in a realistic operative scenario, the robotic arms would be travelling well-defined trajectories allowing for faster safe speeds. This would allow faster update rates and MAMR's speeds.}




\bibliographystyle{IEEEtr}
\bibliography{references}

\end{document}